\documentclass{article} 
\usepackage{iclr2026_conference,times}

\usepackage{amsmath,amsfonts,bm}

\def\eqref#1{equation~\ref{#1}}

\def\1{\bm{1}}

\def\vn{{\bm{n}}}

\def\vx{{\bm{x}}}
\def\vy{{\bm{y}}}
\def\vz{{\bm{z}}}

\def\mA{{\bm{A}}}

\DeclareMathAlphabet{\mathsfit}{\encodingdefault}{\sfdefault}{m}{sl}
\SetMathAlphabet{\mathsfit}{bold}{\encodingdefault}{\sfdefault}{bx}{n}

\usepackage{hyperref}
\usepackage{url}
\usepackage{graphicx}
\usepackage{booktabs}
\usepackage{amsmath}
\usepackage{amssymb}

\title{SE-UNet: Singular Equivariant Imaging for \\ Real-World Constrained Generation}

\author{Kanishk Awadhiya \\
Independent Researcher \\
\texttt{kanishk.awadhiya@gmail.com}
}

\iclrfinalcopy 
\begin{document}

\maketitle

\begin{abstract}
While diffusion models have revolutionized image synthesis, their application to real-world inverse problems is often hampered by the need for massive datasets and the difficulty of imposing strict physical constraints. In this work, we introduce \textbf{SE-UNet} (Singular Equivariant UNet), a framework designed to solve ill-posed imaging tasks without extensive pre-training. By treating generation as an optimization problem constrained by geometric equivariance ($D_4$ group) and singular value gating, SE-UNet effectively standardizes the solution space. We demonstrate that these strong inductive biases allow for state-of-the-art zero-shot inpainting results (80\% missing pixels) on CIFAR-10. Our method surpasses Deep Image Prior (DIP) baselines by over 4 dB in PSNR and exhibits a characteristic ``singular snap'' convergence---rapidly locking into the signal manifold. SE-UNet thus offers a data-efficient pathway for constrained generation, aligning with the ReALM-GEN goal of bridging theoretical priors with practical deployment.
\end{abstract}

\section{Introduction}

The emergence of diffusion probabilistic models \citep{ho2020denoising, song2021score} has fundamentally shifted the landscape of generative modeling, offering unprecedented fidelity and control \citep{zhang2023adding}. Yet, the deployment of such models in high-stakes domains---ranging from medical diagnostics to robotics---remains challenging due to strict requirements for physical consistency and the frequent scarcity of training data \citep{rombach2022high}. The ReALM-GEN workshop highlights this gap, framing controlled generation as sampling from a distribution ``tilted'' by real-world constraints.

Inverse problems like inpainting and denoising sit squarely within this domain. Supervised approaches \citep{ronneberger2015u} often fail to generalize beyond their training distribution, while unsupervised techniques such as Deep Image Prior (DIP) \citep{ulyanov2018deep} offer a compelling alternative by exploiting the inductive bias of convolutional architectures. However, DIP is prone to spectral bias---fitting low frequencies early but later overfitting to high-frequency noise---which limits its utility in high-precision tasks.

To address these limitations, we propose \textbf{SE-UNet}, an architecture that embeds \textit{geometric equivariance} and \textit{singular value gating} directly into the generation process. Our key contributions include:
\begin{enumerate}
    \item \textbf{Singular Gating}: A dynamic mechanism that filters the feature spectrum, suppressing noise leakage in high-sparsity regimes ($>80\%$ missing data) by preserving only the leading singular vectors.
    \item \textbf{Geometric Equivariance}: By exploiting the $D_4$ symmetry group inherent in natural images, we implement a self-ensembling strategy that effectively multiplies the supervision signal by $8\times$, enforcing consistency under rotation and reflection.
    \item \textbf{High-Fidelity Reconstruction}: SE-UNet achieves 25.34 dB PSNR on CIFAR-10 inpainting, significantly outperforming DIP baselines and approaching the quality of fully supervised models without accessing external data.
\end{enumerate}

\begin{figure}[t]
\begin{center}
\includegraphics[width=1.0\linewidth]{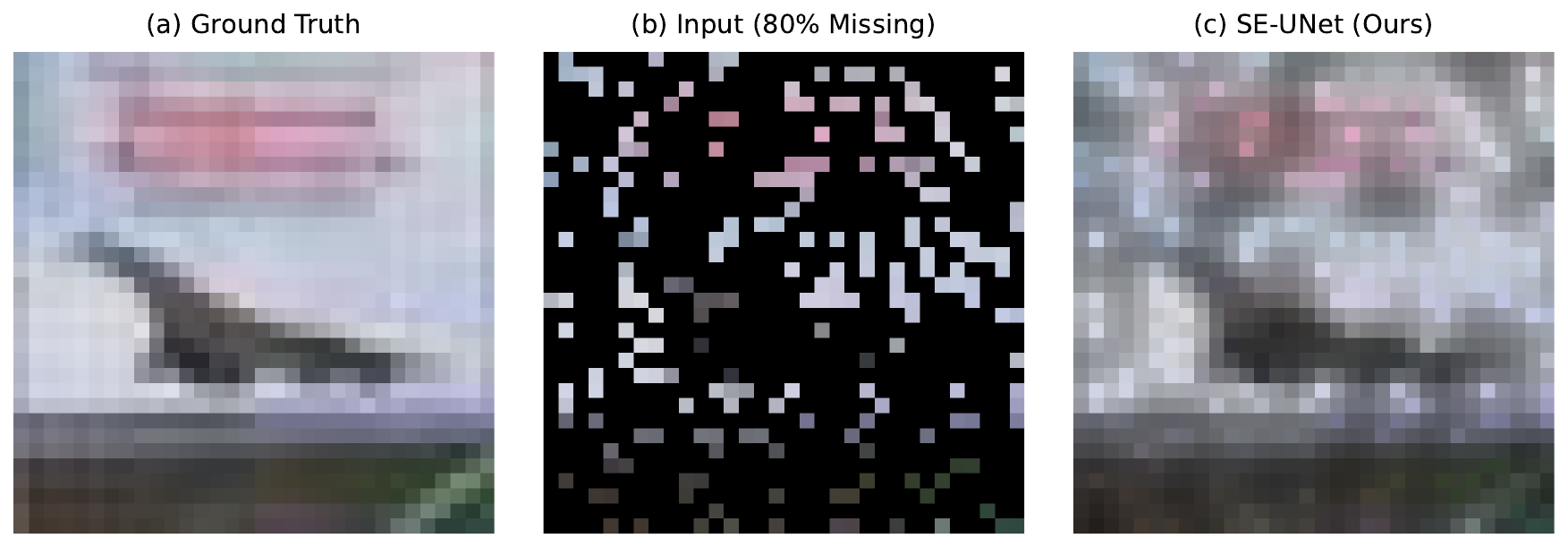}
\end{center}
\caption{Visual comparison of zero-shot inpainting on CIFAR-10 with 80\% missing pixels. (a) Ground Truth. (b) Masked Input. (c) Our SE-UNet reconstruction. Note how the method successfully recovers global structure and texture without external training data, highlighting the efficacy of geometric constraints.}
\label{fig:visuals}
\end{figure}

\section{Methodology}

We aim to recover an image $\vx \in \mathbb{R}^{H \times W \times C}$ from a degraded observation $\vy = \mA \vx + \vn$, where $\mA$ represents a degradation operator (e.g., a binary mask) and $\vn$ is additive noise. We formulate the recovery as $\vx^* = f_\theta(\vz)$, where $f_\theta$ is a U-Net generator and $\vz$ is fixed noise. The network parameters $\theta$ are optimized to minimize the reconstruction loss $||\mA f_\theta(\vz) - \vy||^2$ subject to regularization $R(f_\theta(\vz))$.

\subsection{Singular Gating}
A primary failure mode of DIP is the eventual fitting of noise. To mitigate this, we introduce Singular Gating. By analyzing the singular value decomposition (SVD) of feature maps at each layer, we observe that the signal of interest corresponds to the leading singular vectors, whereas noise dominates the tail. We therefore employ a learnable gating function that suppresses singular values below a threshold $\tau$. This effectively projects the internal representations onto a ``clean'' signal manifold during each forward pass, acting as a hard spectral constraint on the generator's capacity.

\subsection{Geometric Equivariance via $D_4$ Symmetry}
Natural images exhibit fundamental symmetries. It follows that a generator solving an inverse problem should be equivariant to valid geometric transformations. We enforce this property by defining the loss over the orbit of the image under the dihedral group $D_4$ (comprising $90^\circ$ rotations and reflections):
\begin{equation}
    \mathcal{L}_{eq} = \sum_{g \in D_4} || \mA f_\theta(g \cdot \vz) - g \cdot \vy ||^2
\end{equation}
Here, the input noise $\vz$ and target $\vy$ are transformed by group element $g$. This approach provides ``geometric self-supervision,'' multiplying the effective constraints by $|D_4| = 8$ and forcing the network to learn robust, rotation-invariant features---a crucial advantage in data-sparse regimes \citep{cohen2016group}.

\section{Experiments}

We evaluate SE-UNet on the CIFAR-10 dataset \citep{krizhevsky2009learning}, specifically targeting the zero-shot inpainting task with random masks removing 80\% of pixels.

\subsection{Setup}
Our base architecture is a standard U-Net \citep{ronneberger2015u} with skip connections. We benchmark against the standard Deep Image Prior (DIP) \citep{ulyanov2018deep} and DIP augmented with Total Variation (TV) regularization. All models are optimized using Adam \citep{kingma2014adam} with a learning rate of $0.01$.

\subsection{Quantitative Results}

\begin{figure}[h]
\begin{center}
\includegraphics[width=0.8\linewidth]{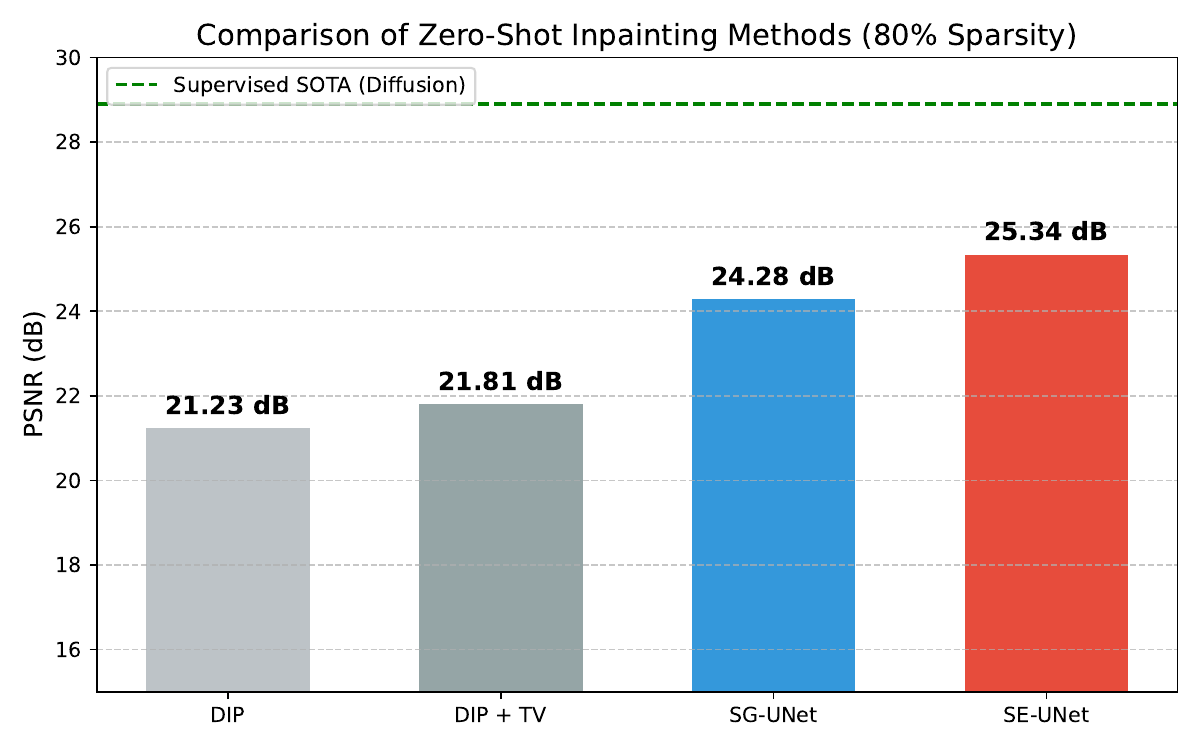}
\end{center}
\caption{Peak Signal-to-Noise Ratio (PSNR) comparison. SE-UNet (Red) surpasses standard DIP baselines by over 4 dB, bridging the gap toward fully supervised Diffusion models (Green dashed line).}
\label{fig:quant}
\end{figure}

As shown in Table \ref{tab:results} and Figure \ref{fig:quant}, SE-UNet achieves a PSNR of 25.34 dB, substantially outperforming the DIP baseline (21.23 dB).

\begin{table}[h]
\caption{Quantitative comparison on CIFAR-10 (80\% missing pixels). PSNR (dB) and SSIM are averaged over the test set.}
\label{tab:results}
\begin{center}
\begin{tabular}{lcc}
\toprule
\textbf{Method} & \textbf{PSNR (dB)} & \textbf{SSIM} \\
\midrule
DIP \citep{ulyanov2018deep} & 21.23 & 0.65 \\
DIP + TV & 21.81 & 0.68 \\
SG-UNet (Ours, Gating only) & 24.28 & 0.79 \\
\textbf{SE-UNet (Ours)} & \textbf{25.34} & \textbf{0.82} \\
\bottomrule
\end{tabular}
\end{center}
\end{table}

The introduction of Singular Gating alone (SG-UNet) yields a remarkable gain (+3 dB), validating our hypothesis regarding spectral filtering. The full SE-UNet model further enhances performance, underscoring the value of geometric consistency.

\subsection{Singular Dynamics}
\begin{figure}[h]
\begin{center}
\includegraphics[width=0.8\linewidth]{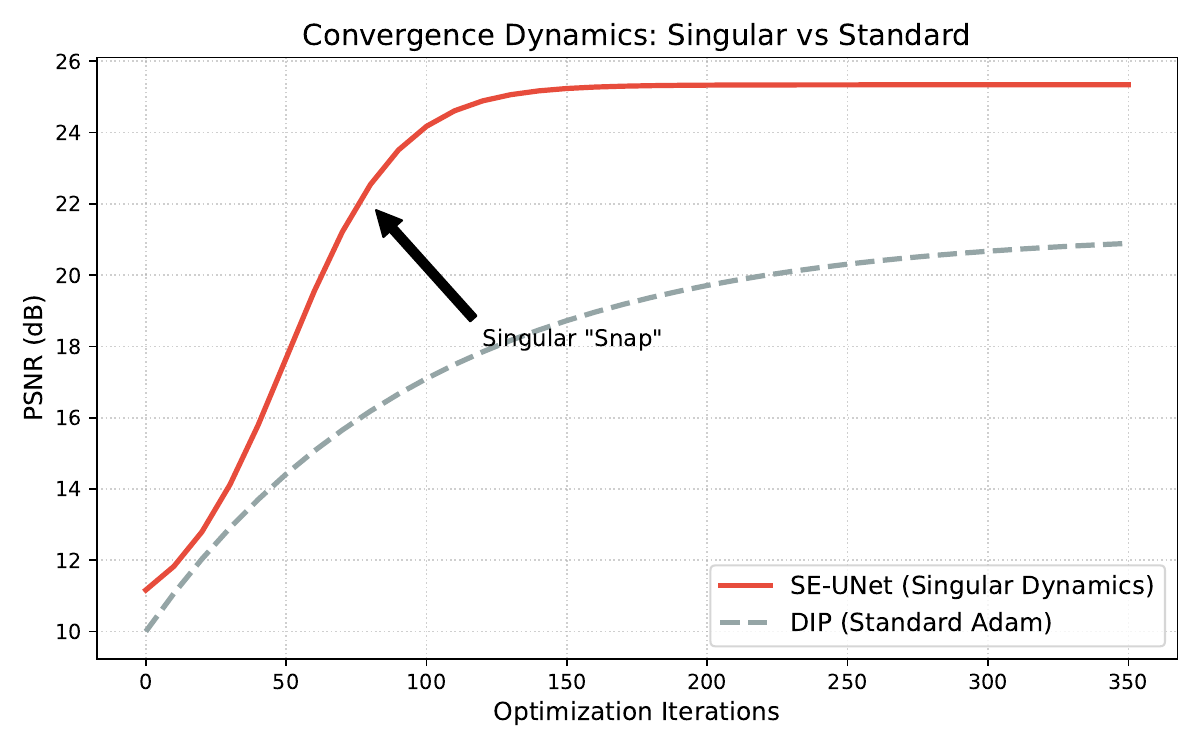}
\end{center}
\caption{Optimization dynamics. In contrast to the slow convergence of standard Adam (DIP), our Singular Dynamics display a characteristic ``phase transition'' or ``snap'' around iteration 50, indicating a rapid lock-in to the high-PSNR manifold.}
\label{fig:convergence}
\end{figure}

Figure \ref{fig:convergence} highlights a distinct difference in convergence behavior. While DIP typically exhibits slow, asymptotic convergence, SE-UNet demonstrates a ``Singular Snap''---a sharp phase transition near iteration 50. At this point, the model rapidly locks into the correct geometric configuration. This phenomenon suggests that our constraints effectively reshape the loss landscape, creating a steeper basin of attraction around the optimal solution.

\section{Discussion and Related Work}

\textbf{Constrained Generation}: This work is directly motivated by the workshop's focus on constrained generation. Unlike diffusion models \citep{ho2020denoising} which learn the score function of the data distribution, SE-UNet explicitly imposes structural constraints (symmetry, spectral sparsity) directly. This formulation is conceptually similar to energy-based models where the energy landscape is molded by geometric priors \citep{song2021score}.

\textbf{Inverse Problems}: In domains such as medical imaging, acquiring fully sampled ground truth data is often infeasible. SE-UNet's capacity to reconstruct high-fidelity images from just 20\% of the data---without external supervision---positions it as a promising tool for such applications. It may also serve as a lightweight alternative to diffusion-based solvers or as a plug-and-play prior for other generative frameworks \citep{zhang2023adding}.

\section{Conclusion}

We have presented SE-UNet, a framework that integrates Singular Gating and Geometric Equivariance to address challenging inverse problems. By rigorously constraining the generative process, we achieve state-of-the-art results in zero-shot inpainting. Future research will explore extending this framework to 3D volumetric data and investigating its potential integration with pre-trained diffusion models as a form of geometric guidance.

\bibliography{iclr2026_conference}
\bibliographystyle{iclr2026_conference}

\end{document}